\UseRawInputEncoding
\documentclass[a4paper,10pt,oneside]{article}
\usepackage[utf8]{inputenc}
\usepackage{icad2023,amsmath,epsfig,times,url,hyperref}
\usepackage[T1]{fontenc}
\usepackage{amssymb}
\usepackage{booktabs}
\usepackage{threeparttable}

\usepackage{amsfonts,amssymb}
\usepackage{amsmath,amsfonts}
\usepackage{algorithmic}
\usepackage{algorithm}
\usepackage{array}
\usepackage[caption=false,font=normalsize,labelfont=sf,textfont=sf]{subfig}
\usepackage{textcomp}
\usepackage{stfloats}
\usepackage{url}
\usepackage{verbatim}
\usepackage{graphicx}
\usepackage{geometry}
\title{Textual-Knowledge-Guided Numerical Feature Discovery Method \\for Power Demand Forecasting}

\twoauthors{Zifan Ning} {Hunan University \\ Changsha, China  \\ {\tt zifan.ning@kcl.ac.uk}}
{Min Jin} {Hunan University \\ Changsha, China  \\ {\tt jinmin@hnu.edu.cn}}

\begin{document}
\ninept
\maketitle
\begin{sloppy}
\begin{abstract}
Power demand forecasting is a crucial and challenging task for new power system and integrated energy system. However, as public feature databases and the theoretical mechanism of power demand changes are unavailable, the known features of power demand fluctuation are much limited. Recently, multimodal learning approaches have shown great vitality in machine learning and AIGC. In this paper, we interact two modal data and propose a textual-knowledge-guided numerical feature discovery (TKNFD) method for short-term power demand forecasting. TKNFD extensively accumulates qualitative textual knowledge, expands it into a candidate feature-type set, collects numerical data of these features, and eventually builds four-dimensional multivariate source-tracking databases (4DM-STDs). Next, TKNFD presents a two-level quantitative feature identification strategy independent of forecasting models, finds 43-48 features, and systematically analyses feature contribution and dependency correlation. Benchmark experiments in two different regions around the world demonstrate that the forecasting accuracy of TKNFD-discovered features reliably outperforms that of SoTA feature schemes by 16.84\% to 36.36\% MAPE. In particular, TKNFD reveals many unknown features, especially several dominant features in the unknown energy and astronomical dimensions , which extend the knowledge on the origin of strong randomness and non-linearity in power demand fluctuation. Besides, 4DM-STDs can serve as public baseline databases.

\bf{Multimodal learning, Qualitative knowledge, Quantitative identification, Feature corpus, DCW algorithm, Integrated energy dimension, Methane price.}
\end{abstract}

\section{Introduction}

\hspace{16pt}Power sectors are currently undergoing fundamental transitions towards sustainability and decarbonization, leading to a tremendous increase in the complexity of power grids. Accurate forecasting of power demand is crucial for planning and dispatching these complex power grids, as well as for supporting policy-making and decision discussions in the competitive context of the integrated energy system (IES) market. However, with the frequent occurrence of extreme climates, the development of various energy storage complementary technologies such as batteries and compressed air, and the emergence of new business models including demand-side management and large-scale aggregation in IES market\cite{asai2020forecasting, yang2019operational}, the nonlinearity and randomness of power demand changes are significantly enhanced, which makes accurate short term power demand forecasting (SPDF) much challenging research. 

\begin{figure*}[!t]
	\centering
	\includegraphics[width=\textwidth]{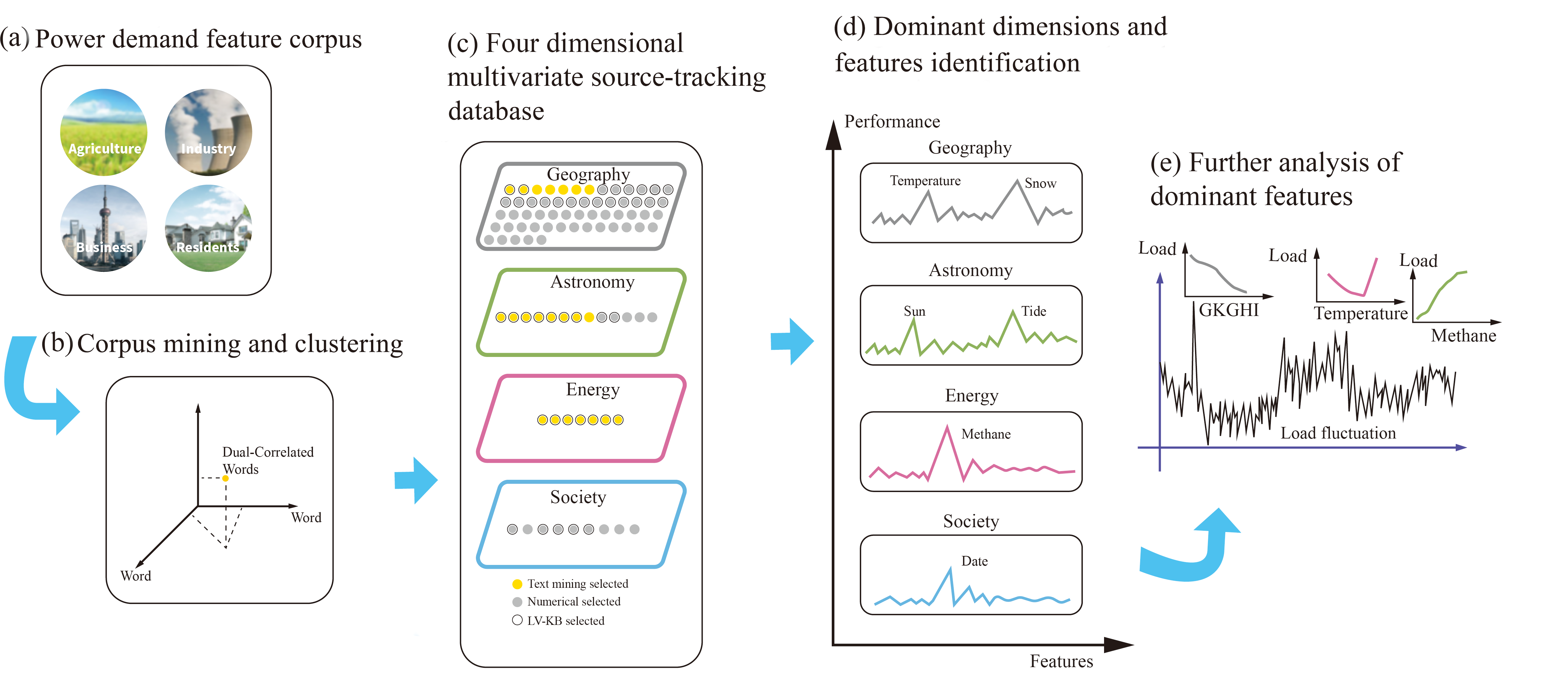}
	\caption{Framework of TKNFD method.}
	\label{fig1}
\end{figure*}

It is well known that data and features determine the upper limit of machine learning, and models and algorithms are only approximating this upper limit. As forecasting models become more sophisticated, research on short-term power demand forecasting (SPDF) has shifted from solely examining power demand to tracking the sources of power load changes and exploring the underlying patterns and features. This shift in focus is particularly relevant for new power systems, where improving forecasting accuracy is crucial \cite{zhu2022review, ye2019data}.  Feature engineering \cite{hafeez2021novel} is an efficient and promising approach for the discovery of influencing factors on power demand change. The feature engineering techniques involve feature construction, feature selection, and feature extraction.

In feature construction, weather is considered the most critical factor affecting power load change \cite{xie2016temperature, xie2016relative, you2022digital}. Lots of research works selected weather-related factors like temperature, humidity, precipitation etc. as candidate features. For example, Son. H et al. \cite{son2017short} studied the influence of moving average temperature and hysteresis temperature of different lengths on SPDF accuracy and then proposed the power load neighbor effect. However, research has shown that weather is not the only factor related to power load fluctuation. According to the features of holiday power load fluctuation, P. Zeng et al. \cite{zeng2019learning} believed that the power load fluctuation law of holidays is entirely different from that of non-holidays, and the historical data onto holidays is relatively sparse. Thus, this study proposed a transfer learning model from transfer learning data onto neighborhood cities. Ernesto et al. \cite{aguilar2021short} train models in specific situations, such as vacations, and weekends, to enhance the accuracy of SPDF during special event days. They found that holiday data has a strong natural randomness, posing an outstanding influence to SPDF. Ruijin Zhu et al. \cite{zhu2019short} considered the correlation between heating, gas and power loads and designed a novel hybrid neural network for SPDF, shows that the coupling between gas load and power load is strong and gas load data improves the forecasting power demand accuracy. Wang et al. \cite{wang2020multi} considered temporal dynamic and coupling characteristics of load series and constructed an extended feature set. The mechanism behind power load fluctuations remains unclear, leading to a lack of theoretical guidance for systematically exploring the factors that influence load fluctuations and their correlations. As a result, existing studies have relied on experiential knowledge to consider one or several influencing factors, such as weather, calendar, social or economic indicators, as features for forecasting. However, the total number of such empirical features remains limited to date.

Feature selection is a crucial step in machine learning, and there are three commonly used methods: filter, wrapper, and embedding. Filter method \cite{zhou2017holographic} determines the importance of features through some statistical indicators such as variance, chi-square test, correlation coefficient and mutual information. The wrapper method \cite{abedinia2016new} involves a search algorithm and a machine learning model that evaluate candidate feature subsets based on model performance. Unlike the previous two methods, embedding method \cite{hafeez2021novel} integrates the feature selection process into the learning algorithm so that feature selection and forecasting are performed simultaneously.  These feature selection methods are all based on the selection of existing features derived from experience. In feature extraction, further features are generated by data-driven deep learning techniques to reflect implicitly complex correlation of multivariate load series \cite{elahe2022knowledge}. At present, there are no publicly available databases for calculating and inferring the correlation between numerous potential influencing factors and power demand, and occasions or scenarios offering limited data sources are more and more common. The applications of feature engineering, such as data-driven feature extraction and feature selection, are far from mature \cite{zhu2022review}.

Recently, multimodal learning has emerged as a promising research direction in machine learning and artificial intelligence, with wide-ranging applications in fields such as computer vision \cite{bayoudh2021survey}, medical diagnosis \cite{lipkova2022artificial}, and AI-generated content \cite{efimova2022text}. Multimodal learning includes multimodal fusion and multimodal interaction. Multimodal fusion \cite{dai2023analysis} aims to integrate multiple sources of data from different modalities, such as audio, video, text, and others, to capture the diversity and completeness of multimodal data. This involves extracting features from each modality using techniques such as convolutional neural networks for images, spectrum analysis for audio, and bag-of-words models for text. The features are then aligned and combined to form a single representation that is used for machine learning tasks such as classification. Multimodal interaction \cite{xiao2022complementary} refers to the process of different modal features influencing and affecting each other. For example, image features can be used to influence text features and vice versa, to better capture the complex relationships and relevance of multimodal data. We believe that both multimodal fusion and interaction are based on the dual attributes of association and complementarity between different modal data. By combining or interacting different modalities, we can construct datasets that are suitable for machine learning models in terms of granularity, scope, and temporality.

Inspired by this, in the field of feature engineering for power demand forecasting, we select two modal data, text and numerical, for one-way interaction, and propose a textual-knowledge-guided numerical feature discovery (TKNFD) method for the first time. TKNFD rapidly accumulates qualitative knowledge of text-based features distributed over the internet, clusters and expands them into a candidate feature-type set, and then collects corresponding numerical data under the guidance of this set. This data is used to build four-dimensional multivariate source-tracking databases (4DM-STD). Based on this database, TKNFD presents a two-level quantitative feature identification strategy that is independent of forecasting models, resulting in the discovery of 43-48 features. Our goal is to use text modal data, which provides coarse-grained, large-coverage scope, and non-temporal qualitative knowledge, to guide the discovery and identification of corresponding numerical modal data, which provides fine-grained, specific forecasting scope, and temporal quantitative features. This text modal knowledge of power demand change is scattered across various countries and regions around the world and has not yet been systematically collected and utilized in the field of power demand forecasting.

The framework of TKNFD is illustrated in Figure 1 as follows. (a) By crawling the text reports of the power system official websites extensively on the power consumption behaviors of four target user groups (agriculture, industry, business, and residents), the first power demand feature corpus is constructed. (b) By mining the proposed corpus, three-domain dimensions (integrated energy, astronomy, and geography) along with their partial features affecting load change are obtained. (c) Guided by the proposed corpus and the research works of our team in numerical feature extraction, we expand the three domain dimensions and construct the four-dimensional multivariate data. (d) The four-dimensional multivariate data and historical load data constitute the four-dimensional multivariate source-tracking database (4DM-STD) as the numerical candidate feature database. Hierarchical identification of dominant dimensions and features that impact load fluctuation are then performed on the feature database. (e) The contribution of different features to power demand fluctuation and the dependence correlation between them are further systematically analyzed. 

The rest of the paper is organized as follows. Section 2 describes the discovery of a candidate feature-type set guided by textual knowledge. Section 3 presents the construction of the feature database 4DM-STD and the identification of dominant dimensions and features. Benchmark forecast experiments on two different regions are demonstrated to compare the performance of TKNFD found features with that of the state-of-the-art feature schemes in Section 4. Section 5 further analyzes the contribution and the dependence correlation. Section 6 concludes this study.

\section{Discovery of Candidate Feature Set}
\subsection{\textit{Construction of Power Demand Feature Corpus}}

\hspace{16pt}Electricity users are commonly classified into four categories: agriculture, industry, business, and residential. To comprehensively gather knowledge on the various features affecting power consumption changes, this study uses these four categories of users as an initial clue. We crawl over a thousand text reports on the four user groups released by the U.S. energy government website$\footnote{https://www.energy.gov/}$, and a power demand feature corpus is rapidly constructed.

As natural language processing continues to advance \cite{mcenery2019corpus, wang2021investigating}, there is increasing interest in using corpus analysis \cite{ardila2019common, williams2017broad} to discover text similarities. The mutual information algorithm \cite{yang2022learning, bachman2019learning} has proven effective in measuring similarity relationships between discrete variables, and it can also be used for correlation analysis between continuous variables. Mutual information does not require categorical ordering and is derived from the joint probability distribution of random variables $X$ and $Y$, as well as their marginal probability distributions $p(x)$ and $p(y)$.

\begin{equation}
	I(X; Y) = \sum_{y\in \mathcal{Y}}^{ }\sum_{x\in \mathcal{X}}^{ } p(x, y) log(\frac{p(x, y)}{p(x)p(y)})
\end{equation}

However, experiments show that only focusing on exploring the similarity between texts algorithm cannot achieve ideal results in exploring the dominant factors affecting SPDF, due to the low independence between these similarity matching words negatively affects the result, such as $(load, power)$. These highly similar matching words can not help explore the SPDF-related feature. Therefore, this paper explores the similarity of texts, eliminates words with low independence, and increases the weight of valid words such as $(load, weather)$. However, there are still few ideal algorithms to integrate words’ correlation and independence, which could keep the similarity between words while removing the low independent elements\cite{amado2018research,eskici2018text} We propose dual-correlated words (DCW) algorithm to find words (mainly features) that affect power load fluctuation in the power demand feature corpus, as shown in Formula(2). Similarities between Words in a corpus are represented by cosine values of the angles between word vectors. Based on Mutual Information, the independence of words is represented by the PMI (Pointwise Mutual Information) between texts. 

\begin{equation}
	\begin{aligned}
		PMI(word1, word2) = log\frac{P(word1, word2)}{P(word1)P(word2)}
	\end{aligned}
\end{equation}

PMI finds the co-occurrence of words from the perspective of statistics to analyze whether there is semantic correlation or thematic correlation between words. In this task, we set Word 1 as "load," and word2 as each after traversing through the article. DCW solves the correlation between the two words through the comprehensive treatment of the similarity and independence of the two words to find the correlation features that affect the SPDF.

\begin{equation}
	\begin{aligned}
		DCW = \frac{PMI(word1, word2)}{cosine(word1, word2)}
	\end{aligned}
\end{equation}

\subsection{\textit{Text mining on power demand feature corpus}}

\hspace{16pt}Assuming we do not have any prior knowledge of the subject, we need TKNFD to process and classify the features we find. We implement the unsupervised K-Means Clustering algorithm with the word vector model built in the previous step to classify the discovered features, as shown below.

\begin{equation}
	\begin{aligned}
		a_k = \frac{\sum_{i=1}^{n}z_{ik}x_{ij}}{\sum_{i=1}^{n}z_{ik}} 
	\end{aligned}
\end{equation}
\begin{equation}
	\begin{aligned}
		z_{ik} = \left\{
		\begin{array}{lr}
			1, if ||x_i - a_k||^2 = min ||x_i - a_k||^2, &  \\
			0, otherwise, &  
		\end{array}
		\right.
	\end{aligned}
\end{equation}

Where $X = {x_1, ..., x_n}$ is the dataset in a D-dimensional Euclidean spac $\mathbb{R}^d$. $A = {a_1, ..., a_c}$ are the c cluster centers. Let $z = [z_{ilk}]_{n*c}$, where $z_{ik}$ is a binary variable indicating if the data point $x_i$ belongs to $k_{th}$ cluster, $k = 1, ..., c$. 

The found features are classified into approximately three dimensions through k-means neighbour clustering algorithm\cite{jakawat2019graph,zhang2019construction}, as shown in Figure 2. In addition to the dimensions that have been fully studied, such as geography and astronomy, a new classification has emerged, which mainly includes influencing factors such as biological gases and fossil fuels. We call this category the integrated energy dimension, a crucial impact dimension that previous studies have overlooked. The DCW weight between the word 'load' and a correlated word is shown in this figure. Mark '—'  represents that the DCW value of a word and the word 'load' is not higher than the set threshold (0.1). Figure 2 shows that in the report on the official website of the US energy government\cite{hughes2019value,dixon2010us}, the influencing factors of power load change mainly involve the domain features of three dimensions: geography, astronomy, and integrated energy. In the geographical dimension, weather and wind-related features influence power load in agriculture, industry, and the economy. In the astronomical dimension, the main features related to the sun significantly affect industrial power consumption; In the integrated energy dimension, the two features of clean energy and fossil energy are highly weighted.

\begin{figure}[!t]
	
	\centering
	\includegraphics[width=3in]{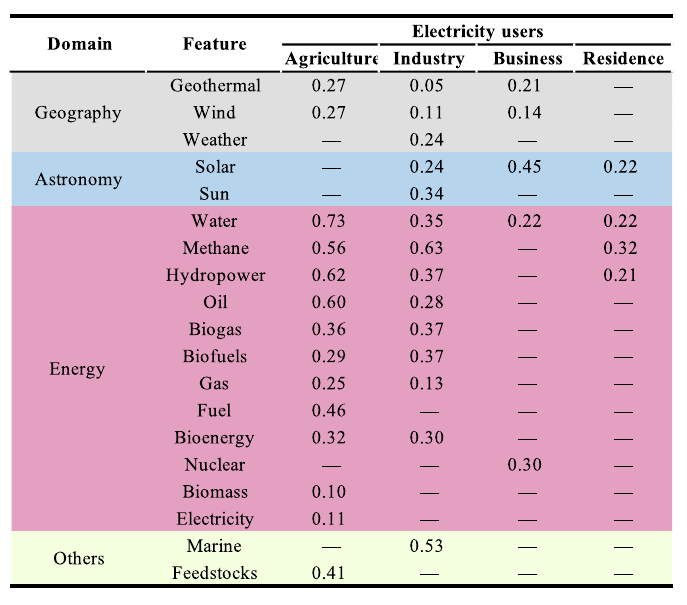}
	\caption{DCW score of power demand feature corpus.}
	\label{fig2}
\end{figure}

\subsection{\textit{Creation of candidate feature-type set}}

\hspace{16pt}We get three main dimensions from power demand feature corpus though previous research. Inspired by the above feature corpus, along with the research work of our team in numerical feature extraction\cite{zhou2017holographic}, we extend the domain features from three dimensions to four dimensions: geography, astronomy, integrated energy, and society. Each dimension consists of multivariate domain feature-types. The geographical dimension includes weather, temperature, air pressure, humidity, etc. The astronomical dimension includes sunshine duration, civil dawn duration, lunar phase, tide, solar radiation, sunrise time, etc. The integrated energy dimension includes methane price, propane price, etc. The social dimension includes holidays, important events, week types, etc.

The four-dimensional multivariate feature-types and historical power load feature-type constitute a candidate feature-type set, as shown in the Formula (6).

\begin{equation}
	X = [G_1, ..., G_i; A_1, ..., A_j; I_1, ..., I_k; S_1, ..., S_l; L_1, ..., L_m]
\end{equation}

$ G, A, I, S, $ and $ L$ respectively represent domain feature dimension of geography, astronomy, integrated energy, society and historical power load. $ i, j, k, l, $ and $ m$ represent the number of domain feature-types in each feature dimension.

With the the candidate feature-type set under the accumulation and inspiration of textual modality knowledge, the dominant dimensions and out-performed features within can be revealed in subsequent studies.

\section{Identification of Dominant Dimensions and Features}

\hspace{16pt}Having obtained the candidate feature-type set through the accumulation and inspiration of textual knowledge, the question arises as to which of the four candidate dimensions has the most significant impact on power demand, apart from the historical demand load dimension. Likewise, it is important to identify the features that critically affect the accuracy of SPDF. To answer these questions, numerical feature databases are constructed, and a two-level quantitative feature identification strategy, independent of forecasting models, is employed to analyze the databases from the dominant dimension to the feature level.

\subsection{\textit{Feature database building}}

\hspace{16pt}Guided by the discovered candidate feature-type set, we then attempted to collect numerical data of these different features to build a four-dimensional multivariate source-tracking database (4DM-STD), which is a numerical modality feature database. However, collecting such feature data, involving multiple fields of astronomy, geography, energy, and society, can be challenging as they are not always readily available or accessible. We put significant efforts into collecting them worldwide and eventually succeeded in establishing a 4DM-STD in Maine and Texas, USA, respectively.

Maine's power load data is downloaded from the ISO-New England website$\footnote{https://www.iso-ne.com/}$. Texas load data is collected from The Power Reliability Council of Texas$\footnote{http://www.ercot.com/}$ and the load data type is daily peak load. Geography data is collected from U.S. Data.GOV$\footnote{https://www.data.gov/}$, Natural Earth Data$\footnote{https://naturalearthdata.com/}$ and United States Geological Survey$\footnote{http://usgs.gov}$. Astronomy data is collected with U.S. NOAA$\footnote{http://gis.ncdc.noaa.gov/}$ and NASA's Global Change Master Directory$\footnote{http://gcmd.nasa.gov/index.html}$. Integrated energy data is collected from U.S. National Renewable Energy Laboratory$\footnote{https://www.nrel.gov/research/data-tools.html}$, U.S. Energy Information Administration$\footnote{https://www.eia.gov/}$,MarketInsider$\footnote{https://markets.businessinsider.com/}$, Maine Govenment $\footnote{https://maine.gov}$ and U.S. DR Power$\footnote{https://egriddata.org/}$. Among them, the Texas natural gas price sold to electric power consumers (dollars per thousand cubic feet) and coal price (delivered to electric power station) are released on a monthly basis. In this study, we consider the natural gas and coal price per day to be linearly distributed, and the missing values are random. Therefore, we subject the dataset to multiple imputation based on Bayesian estimation. We build a 4DM-STD on Maine containing 92 candidate features from 2003 to 2017, and a 4DM-STD on Texas containing 58 candidate features from 2002 to 2019.

\subsection{\textit{Dominant dimension identification}}
\hspace{16pt}Here we adopt one of the model-independent SHAP explainer\cite{parsa2020toward} to identify the dominant dimensions that impact demand forecasting.

The result of SHAP analysis on Maine's 4DM-STD is similar to that on Texas'. As shown in Figure 3, the integrated energy dimension has the most significant impact on demand forecasting, with a mainly negative correlation, which indicates that power demand is in some competition with other integrated energy features. The geographical and astronomical dimensions have a secondary impact in the result. The social dimension has the weakest impact among four dimensions.

\begin{figure}[ht]
	
	\centering
	\includegraphics[width=3in]{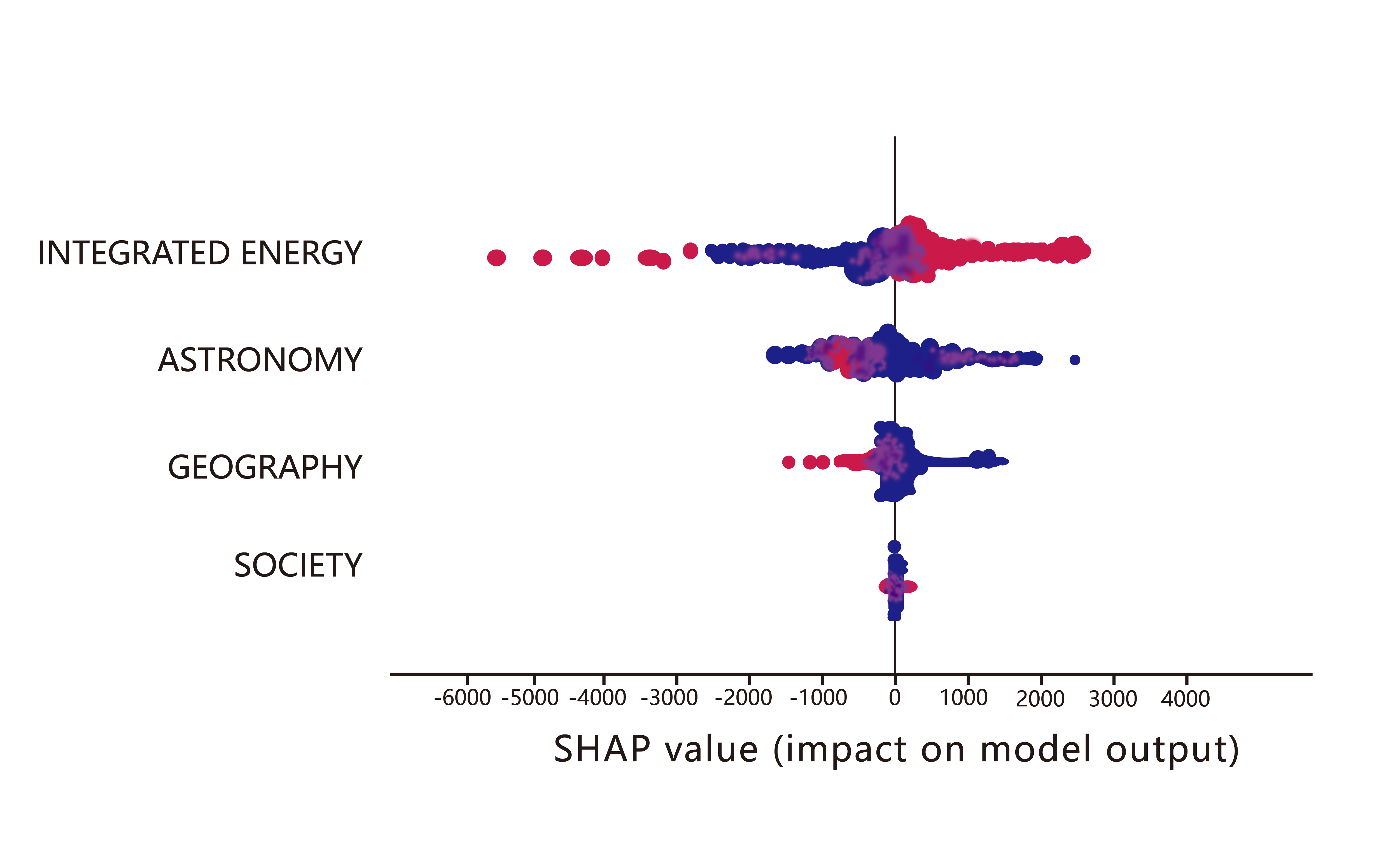}
	\caption{Exploration of dominant dimensions with SHAP analysis of Maine.}
	\label{fig2}
\end{figure}

\subsection{\textit{Feature identification}}

\hspace{16pt}LV-KB, a numerical feature filter algorithm, is chosen for dominant feature identification on 4DM-STD. LV-KB is a combination of Variance Thresholding and Select-K-Best. Variance Thresholding is a fast and lightweight way to eliminate the low variance features that do not express valid information. Using the Variance Threshold as the first feature filter can effectively improve the validity of the dataset and the computational efficiency of the model. According to the research experience of our team, the threshold of variance is set as 0.88. The SelectKBest method, which is a univariate regression function used for further feature filtering. Assuming that the dimension of the candidate feature is n dimensions, the  regression function first calculates the correlation between each feature and the label, as shown in formula (7).

\begin{equation}
	r_i = \frac{\sum_{j=1}^{n}(X_{ij} - \bar{X})(y_j - \bar{y})}{\sqrt{\sum_{j=1}^{n}(X_{ij} - \bar{X})^2\sum_{j=1}^{n}(y_{ij} - \bar{y})^2}}
\end{equation}

Then, the score $f_i $ of the ith feature can be calculated by formula (8).

\begin{equation}
	f_i = \frac{r_i^2}{1 - r_i^2} (n - 2)
\end{equation}

Feature filtering is then carried out according to the ranking of scores. Multiple experimental data show that when the threshold is set to 10, this filter can effectively reject invalid features and retain critical features. Dominant features are selected as shown in Figure 4.

\begin{figure}[!t]
	\centering
	\subfloat[]{\includegraphics[width=2.92in]{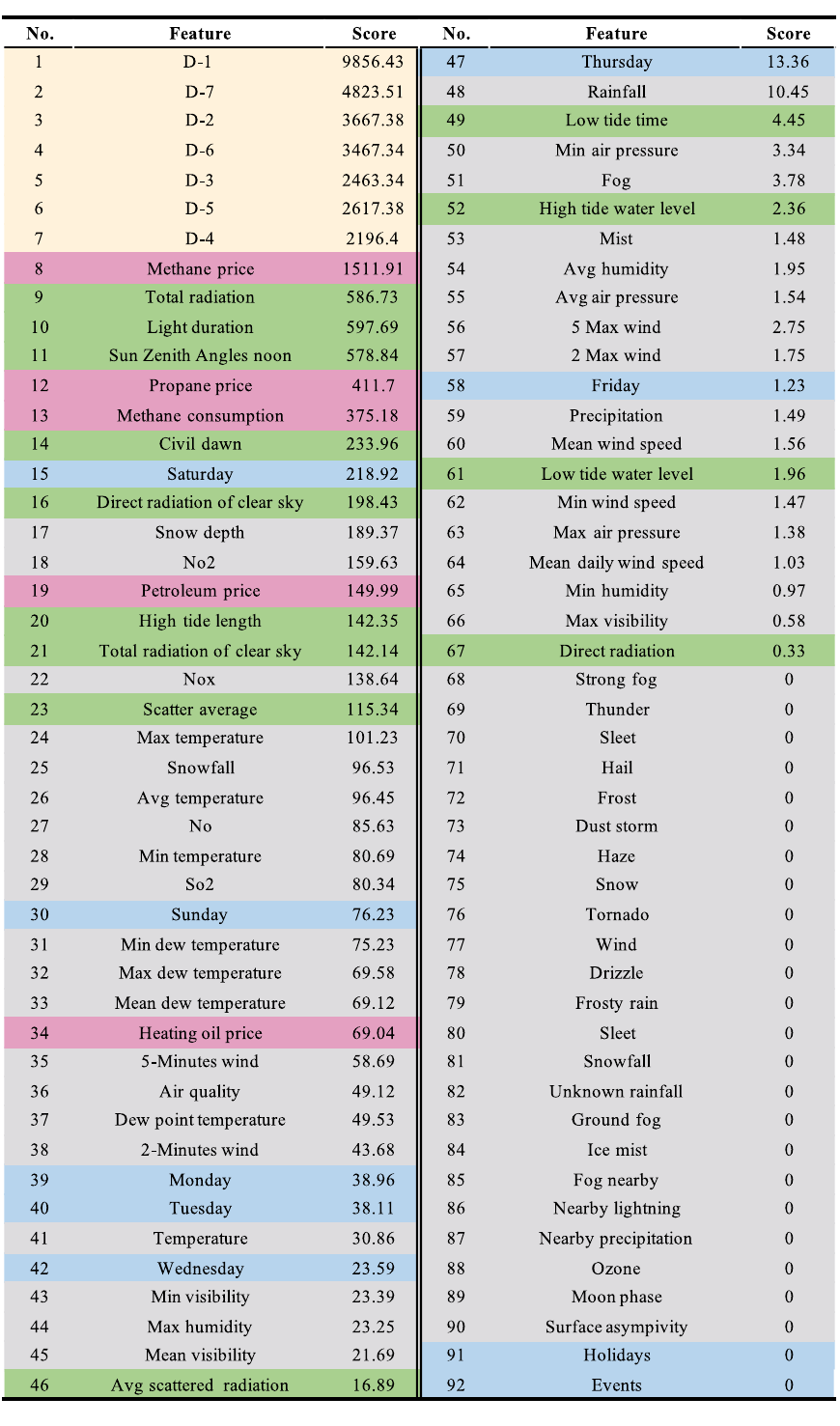}
	}\\
	\subfloat[]{
		\includegraphics[width=2.92in]{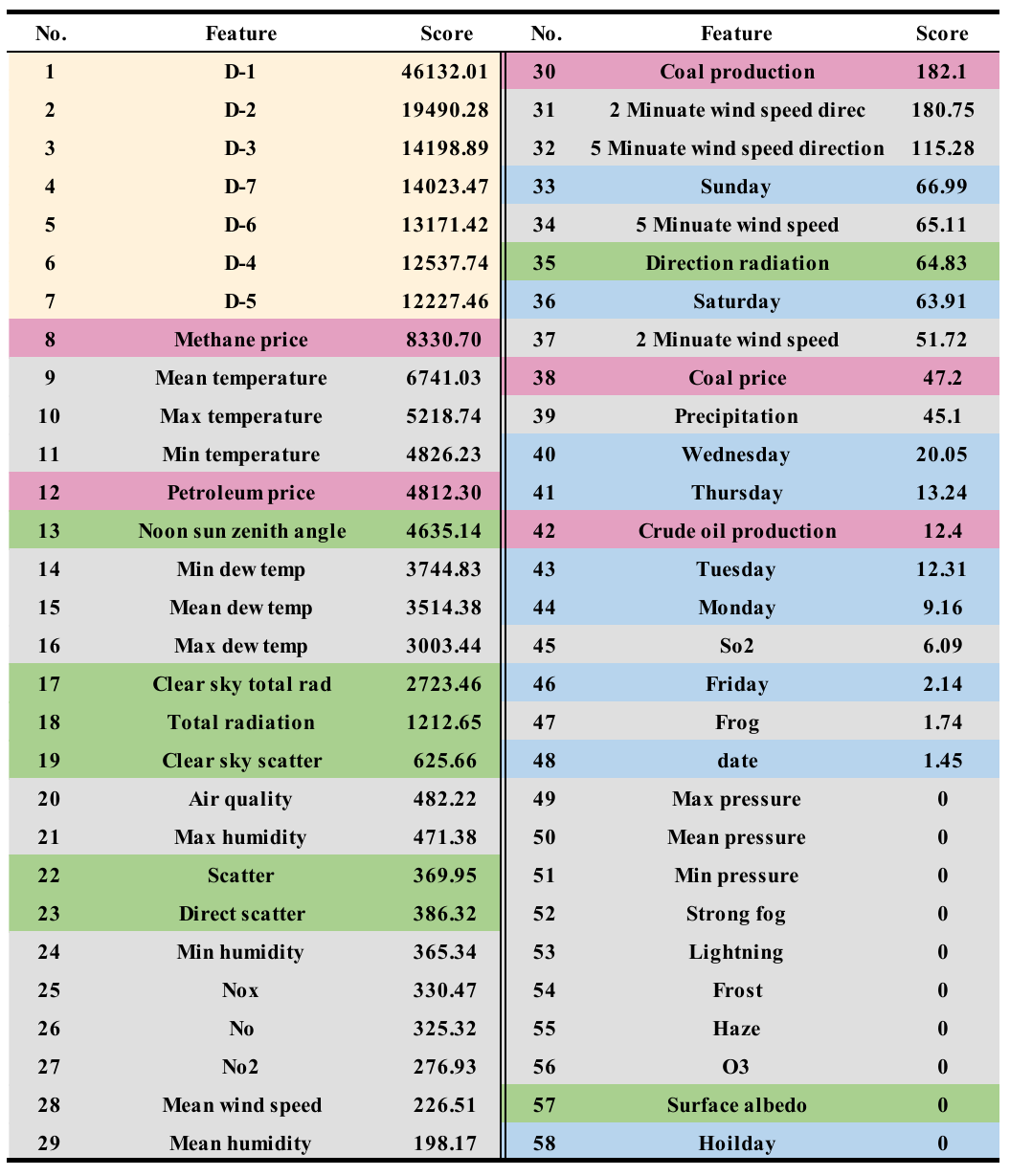}
		}\\

	\includegraphics[width=3in]{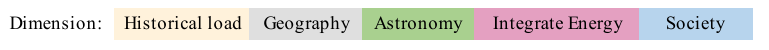}%
	\caption{Dominant feature identification in (a) Maine and (b) Texas.}
	\label{fig10}
\end{figure}

As shown in Figure 4 (a), 48 features are identified in Maine, in which 5 features belong to the integrated energy dimension,  28 to the geographic, 9 to the astronomical, and 6 to the social. As shown in Figure 4 (b), 43 features are identified in Texas, in which 5 features belong to the integrated energy dimension, 26 to the geographic, 7 to the astronomical, and 5 to the social. It is demonstrated in Figure 4 (a) and (b) that the TKNFD-identified 48 features in Maine are similar to the TKNFD-identified 43 features in Texas. In the integrated energy and social dimension, methane price and Saturday get the highest score respectively, indicating that methane price and Saturday are the dominant features in the two dimensions, respectively. In astronomical and geographical dimensions, the sun-related features and temperature have higher scores than other features, implying that the sun-related features and temperature play essential roles in the two dimensions, respectively. In addition, except for the historical load dimension, the features of the integrated energy dimension rank the highest and those of the social dimension rank the lowest. This result is consistent with the conclusion in Section III-B that the integrated energy dimension has the most significant impact while the social dimension has the weakest impact on demand forecasting among the four dimensions.

\section{Forecasting experiments and performance comparison of proposed features}

\subsection{\textit{Forecasting experiments overview}}

\hspace{16pt}In order to investigate the effectiveness of the proposed features found by TKNFD, this paper selects three classical machine learning algorithms (SVR, GBRT, and MLPR) and two typical deep learning models (LSTM\cite{hewamalage2021recurrent}, and Transformer) to build the forecasting models. Besides, transformer model contains totally 52 layers, including convolutional layers, dense layers, extract patch layers, attention layers, LSTM layers and transpose layers.We select MAPE, RMSE, and MAE for performance evaluation. 

In case 1, the data from 2003 to 2014 in Maine is taken as the training set and the data from 2015 to 2017 as the test set. In case 2, the daily peak load from 2002 to 2015 in the Power Reliability Council of Texas region is selected for the training set, and 2006, solstice, 2019 is taken as the testing set.

\subsection{\textit{Forecasting performance of proposed features}}

\hspace{16pt}We apply the data of proposed features found by TKNFD on five models (SVR, GBRT, MLPR, LSTM and Transformer) to forecast the power load in Maine and Texas. Figure 5 presents the experimental result.

As shown in Figure 5, the MAPE of the proposed features is lower than that of the single historical load feature on all five models and in both regions, with a maximum reduction of 42.86$\%$ and a maximum reduction of 40.00$\%$. Especially, the proposed features combined with the transformer model achieved a high accuracy of 1.60$\%$ MAPE in Maine.

In addition, it is also observed that the overall forecasting result of Texas is worse than that of Maine. In the subsequent experiments of these two regions, we will use the better or better forecasting models as benchmark models to further investigate the performance of the proposed features.

\begin{figure}[h]
	\centering
	\subfloat[]{\includegraphics[width=3in]{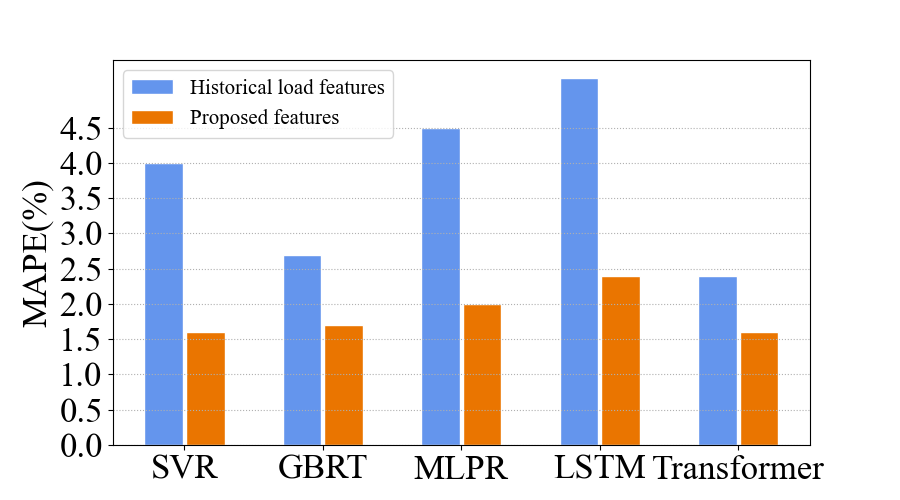}
		\label{fig101}}\\
	\subfloat[]{\includegraphics[width=3in]{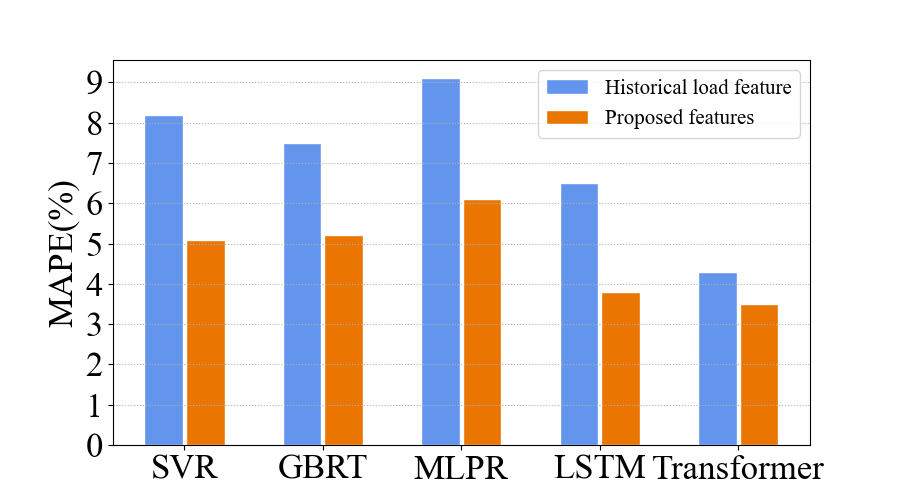}%
		\label{fig102}}
	\caption{Performance of features in (a) Maine and (b) Texas.}
	\label{fig10}
\end{figure}

\subsection{\textit{Comparison with the SoTA feature schemes}}
\subsubsection{Maine}

\hspace{16pt}Based on the SoTA researches, we design five feature schemes for comparison forecasting experiments in Maine. Scheme 1 (S1), which is the initial scheme, only includes date coefficient and the history peak load provided by ISO-New England dataset. All the following cases are built based on S1 ISO-New England dataset. Scheme 2 (S2) adds 3 temperature-related features proposed by\cite{rajbhandari2021impact} on the basis of S1. Scheme 3 (S3), which is the Smart Grid Smart City dataset proposed by\cite{tang2022short} on the basis of S2, filters nine social features exits in the proposed feature database. Scheme 4 (S4) adds three new energy features\cite{zhu2019short} based on S3; Scheme 5 (S5) is the 4DM-STD, which is proposed in this paper and shown in Figure 4. Figure 6 illustrates the designed feature schemes, where "\checkmark" represents the feature is included in this scheme. 

The data from 2003 to 2014 in Maine is taken as the training set and the data from 2015 as the test set. According to the performance results in Figure 5(a), we choose SVR, GBRT and Transformer, three models with better forecasting in Maine, as benchmark forecast models in this comparison experiment. The comparison experimental result is shown in Figure 7.

Figure 7 shows that S1 has the worst forecasting performance, S2 and S3 is close, and the forecasting accuracy of S5 is the highest. S1 only contains seven historical load features and date coefficient, while a large number of studies have shown that temperature is an essential factor affecting load changes. Compared with S1, S2 adds temperature features, and the forecasting accuracy is greatly improved, reflecting the correlation between temperature and load fluctiuation. S3 has nine more features than S2, including holidays, events, and week-types. As can be seen in Section III, society dimension has limited impact on SPDF. Therefore, the improvement of accuracy brought by S3 is less effective. Compared with S3, S4 adds some integrated energy features and astronomy features, and the forecasting accuracy is much improved. S5 entirely covers the four dimensions of geography, society, integrated energy, and astronomy in addition to historical loads, and there are also more features in different dimensions, especially the astronomical features and intergrated energy features. Clearly, S5, the features discovered by TKNFD, shows the best forecasting accuracy among all feature schemes on any forecasting model. The MAPE of S5 is 16.84\% to 36.36\% lower than that of S4, and the MAPE of S5 is less than 1.62\% on all three forecasting models.

\begin{figure}[ht]
	
	\centering
	\includegraphics[width=3in]{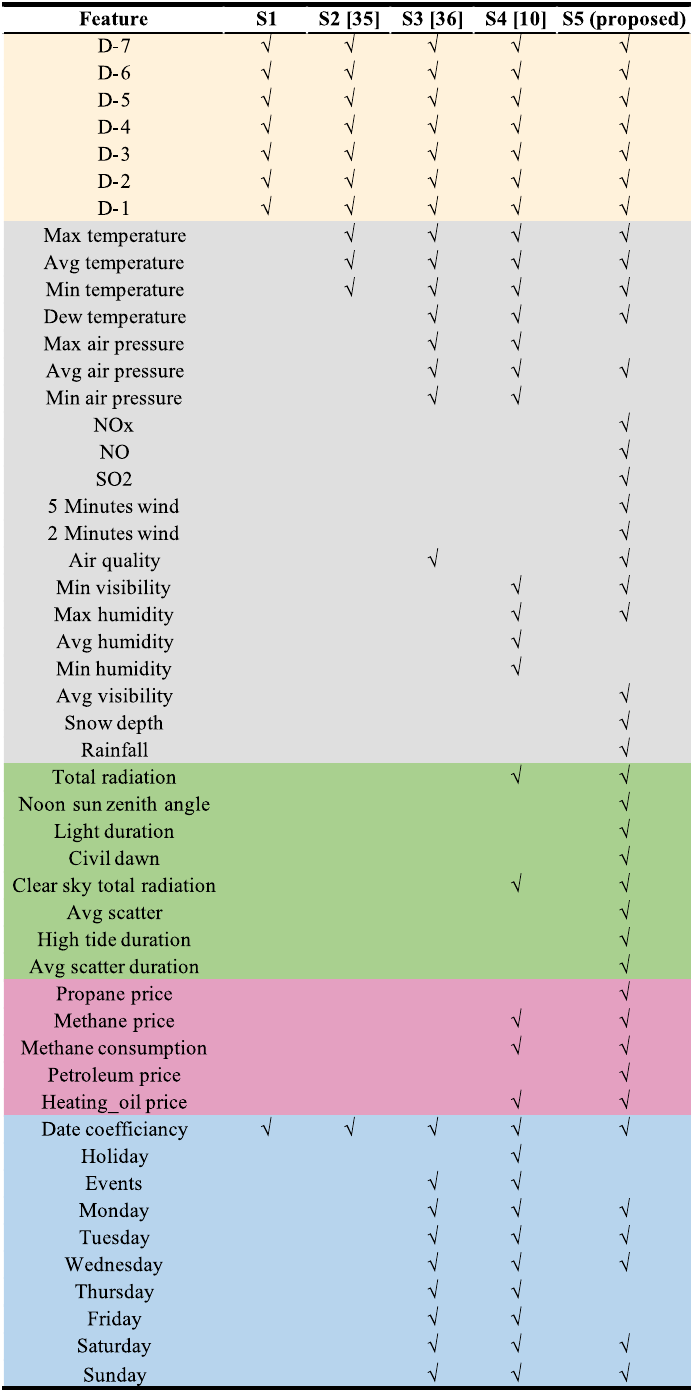}\\
	\includegraphics[width=3in]{pics/instruction}%
	\caption{Feature schemes in Maine.}
	\label{fig2}
\end{figure}

\begin{figure}[!ht]
	
	\centering
	\includegraphics[width=3in]{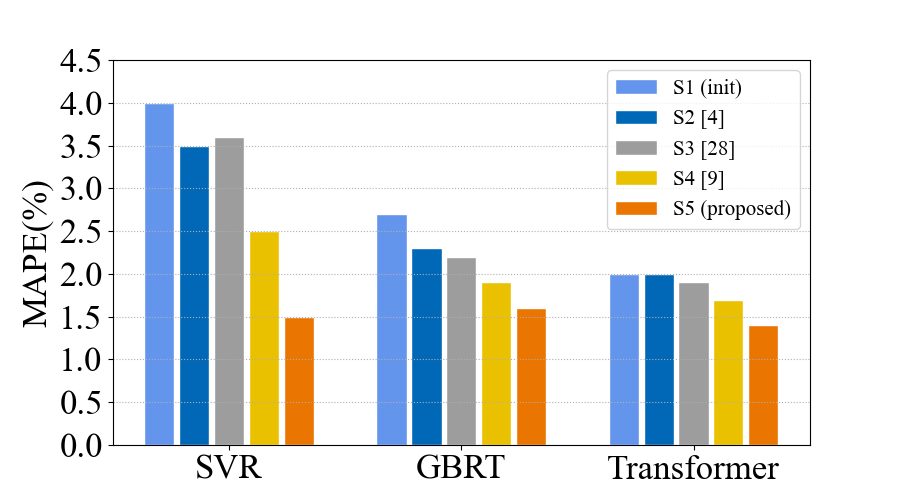}
	\caption{Comparison experimental result in Maine.}
	\label{fig2}
\end{figure}

\subsubsection{Texas}

\hspace{16pt}In Figure 5, five models (SVR, GBRT, MLPR, LSTM and Transformer Model) all show worse forecasting performance in Texas than in Maine. Bayesian Accumulated Regression Tree  (BART) is an algorithm that combines Bayesian theory and the Accumulated Tree model. In the field of demand forecasting, this method can effectively capture the complex relationship between influencing factors and power demand load \cite{alipour2019assessing}. In this comparison experiment in Texas, we take BART as the benchmark forecast model.

The feature scheme in  \cite{alipour2019assessing} includes meteorological factors (such as daily maximum temperature and dew point temperature) and social factors. In contrast, the proposed features discovered by TKNFD consist of not only meteorological factors (belonging to the geographical factors)and social factors, but also lots of integrated energy factors and astronomical factors. 

Table I presents the experimental forecasting result of in Texas. Compared with the feature scheme in \cite{alipour2019assessing}, the proposed features decrease the RMSE and MAE of BART model by 36.6\% and 28.9\%, respectively, proves that the proposed features could significantly improve the SPDF accuracy, even for the advanced forecasting model.

\begin{table}[h]
	\caption{Case 2 Forecasting Result}
	\centering
	\begin{tabular}{c c c c}
		\toprule[1.5pt]
		Features & Model & RMSE & MAE\\
		\midrule[0.75pt]
		Metro+Society\cite{alipour2019assessing} & BART\cite{alipour2019assessing} & 2.866 & 2.213 \\
		The proposed & BART\cite{alipour2019assessing} & 1.817 & 1.572\\
		\bottomrule[1.5pt]
	\end{tabular}
\end{table}

\subsection{\textit{Summary}}

\hspace{16pt}The proposed set of 43-48 features discovered by TKNFD was compared to five feature schemes, which included a single historical load feature and four empirical feature schemes from the state-of-the-art literature. These comparison experiments were conducted using five forecasting models in two different regions. Whether using classical machine learning models or deep learning models, simple models such as SVR or advanced composite models such as BART, whether for the cold northern regions such as Maine or for the hot southern regions such as Texas, the proposed features consistently demonstrated much higher forecasting accuracy than the other five feature schemes. Notably, the proposed features combined with the Transformer model achieved a high accuracy of 1.4\% MAPE in Maine. These results indicate that TKNFD can effectively capture the important source factors affecting load fluctuation, not only in the empirically known geographical and social dimensions but also in the unknown astronomical and integrated energy dimensions. Additionally, each dimension covers more diverse features, making TKNFD a comprehensive and reliable tool for feature discovery in various forecasting models.

\section{Further analysis of proposed features}
\subsection{\textit{Sensitivity analysis of feature contribution}}
\hspace{16pt}Sensitivity analysis refers to studying the uncertainty in the output of a model and further determining the source of uncertainty to study the degree of output change caused by the change of input parameter. Sobol sensitivity analysis is a variance-based sensitivity analysis. It quantifies the uncertainty of the input and output in the form of a probability distribution and decomposes the output variance into parts that can be attributed to the input variables and combinations of variables. Any model may be viewed as a function $Y =f(x)$.
\begin{equation}
	Y = f_0 + \sum_{i=1}^{d}f_iX_i + \sum_{i<j}^{d}f_{ij}(X_j, X_j) +...+f_{1,...,d}(X_1,...,X_d)
\end{equation}
where $f_0$ is a constant and $f_i$ is a function of $X_i$, $f_{ij}$ a function of $X_i$ and $X_j$, etc.

To ensure that the data dimension of the feature does not affect the results of the sensitivity analysis, we normalize all the feature data separately and combine them for a sensitivity analysis in dimension units. The Sobol method is selected to calculate the parameter sensitivity of the MLPRegression model that performs well above. Assuming that the parameters are evenly distributed, the Monte Carlo sample number n takes 1000, and analyzes the four parameters. The results of Sobol sensitivity analysis are shown in Table II. Maine and Table III. Texas.

Where S1 represents the First-order indices, S2 represents the Second-order indices, which reveals the intensity of the interaction between the two parameters. ST represents the Total-effect index. S1conf, S2conf and STconf represent the corresponding confidence level, respectively

The tables show that SPDF is more sensitive to changes in integrated energy parameters and less sensitive to social parameters. The Second-order indices (S2) reveal the intensity of the interaction between two parameters. In this case, S2 analysis shows that the intersection between different dimensions is not vital, and there is no apparent correlation. The Sobol sensitivity analysis confirms the different contributions of the four dimensions identified by SHAP analysis in Section III.

\begin{table}[!h]
	\caption{Maine sobol sensitivity analysis result}
	\centering
	\begin{threeparttable}
		\begin{tabular}{c c c c c c c }
			\toprule[1.5pt]
			Tasks& ST & S1 & S2 & STconf & S1conf & S2conf\\
			\midrule[0.75pt]
			G &  0.1568 & 0.1564 & non & 0.0144 & 0.0347 & non\\
			A & 0.1505 &  0.1531 & non & 0.0130 &0.0291 &non\\
			I & 0.6882 & 0.6882 & non & 0.0450 & 0.0668 & non\\
			S & 0.0011 & 0.0011 & non & 0.0001 & 0.0030 & non\\
			G+A&non &   non  & 0.0013 & non  & non  & 0.0505\\
			G+I & non  & non  & 0.0017 & non  & non & 0.0445 \\
			G+S & non  & non  & 0.0013 & non  & non  & 0.0594 \\
			A+I & non  & non  & -0.0028 & non  & non  & 0.0440\\
			A+S & non  & non  & -0.0048 & non  & non  & 0.0571\\
			I+S & non  & non  & 0.0002 & non  & non  & 0.0058\\
			\bottomrule[1.5pt]
		\end{tabular}
		\begin{tablenotes}
			\footnotesize
			\item[*] G - Geography, A - Astronomy, I - integrated energy, S - Society.
		\end{tablenotes}
		
	\end{threeparttable}
\end{table}

\begin{table}[!h]
	\caption{Texas sobol sensitivity analysis result}
	\centering
	\begin{threeparttable}
		\begin{tabular}{c c c c c c c }
			\toprule[1.5pt]
			Tasks& ST & S1 & S2 & STconf & S1conf & S2conf\\
			\midrule[0.75pt]
			G &  0.4640 & 0.04554 & non & 0.0044 & 0.0188 & non\\
			A & 0.2965 &  0.2948 & non & 0.0278 &0.0478 &non\\
			I & 0.5460 & 0.5437 & non & 0.0393 & 0.0552 & non\\
			S & 0.1128 & 0.1112 & non & 0.0107 & 0.0276 & non\\
			G+A&non &   non  & 0.0054 & non  & non  & 0.0310\\
			G+I & non  & non  & 0.0010 & non  & non & 0.0325 \\
			G+S & non  & non  & -0.0002 & non  & non  & 0.0290 \\
			A+I & non  & non  & 0.0046 & non  & non  & 0.0860\\
			A+S & non  & non  & -0.0003 & non  & non  & 0.0692\\
			I+S & non  & non  & 0.0013 & non  & non  & 0.0840\\
			\bottomrule[1.5pt]
		\end{tabular}
		\begin{tablenotes}
			\footnotesize
			\item[*] G - Geography, A - Astronomy, I - integrated energy, S - Society.
		\end{tablenotes}
		
	\end{threeparttable}
\end{table}

\begin{figure*}[!h]
	\centering
	\subfloat[]{\includegraphics[width=0.47\textwidth,height=0.44\textwidth]{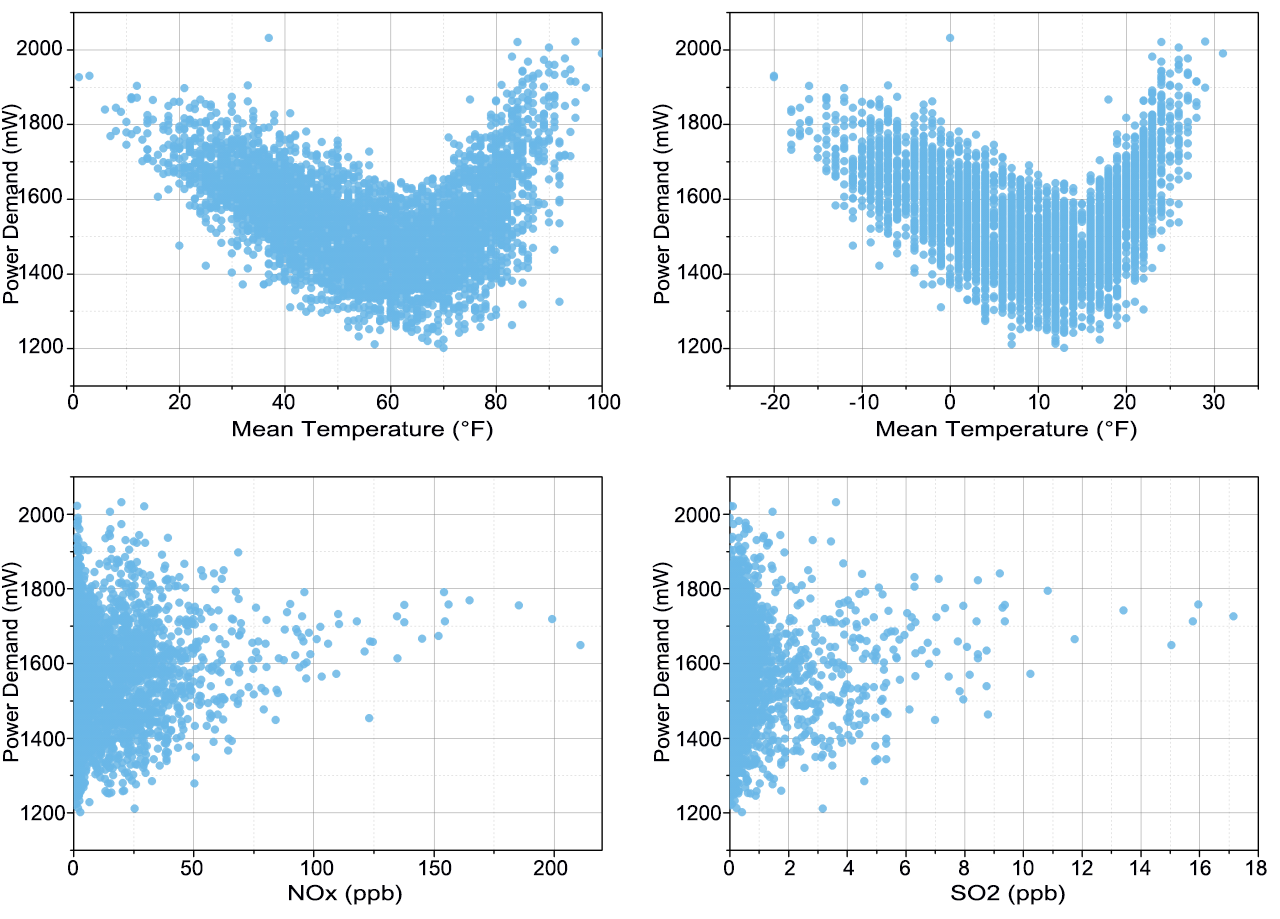}%
		\label{fig71}}
	\hfil
	\subfloat[]{\includegraphics[width=0.47\textwidth,height=0.44\textwidth]{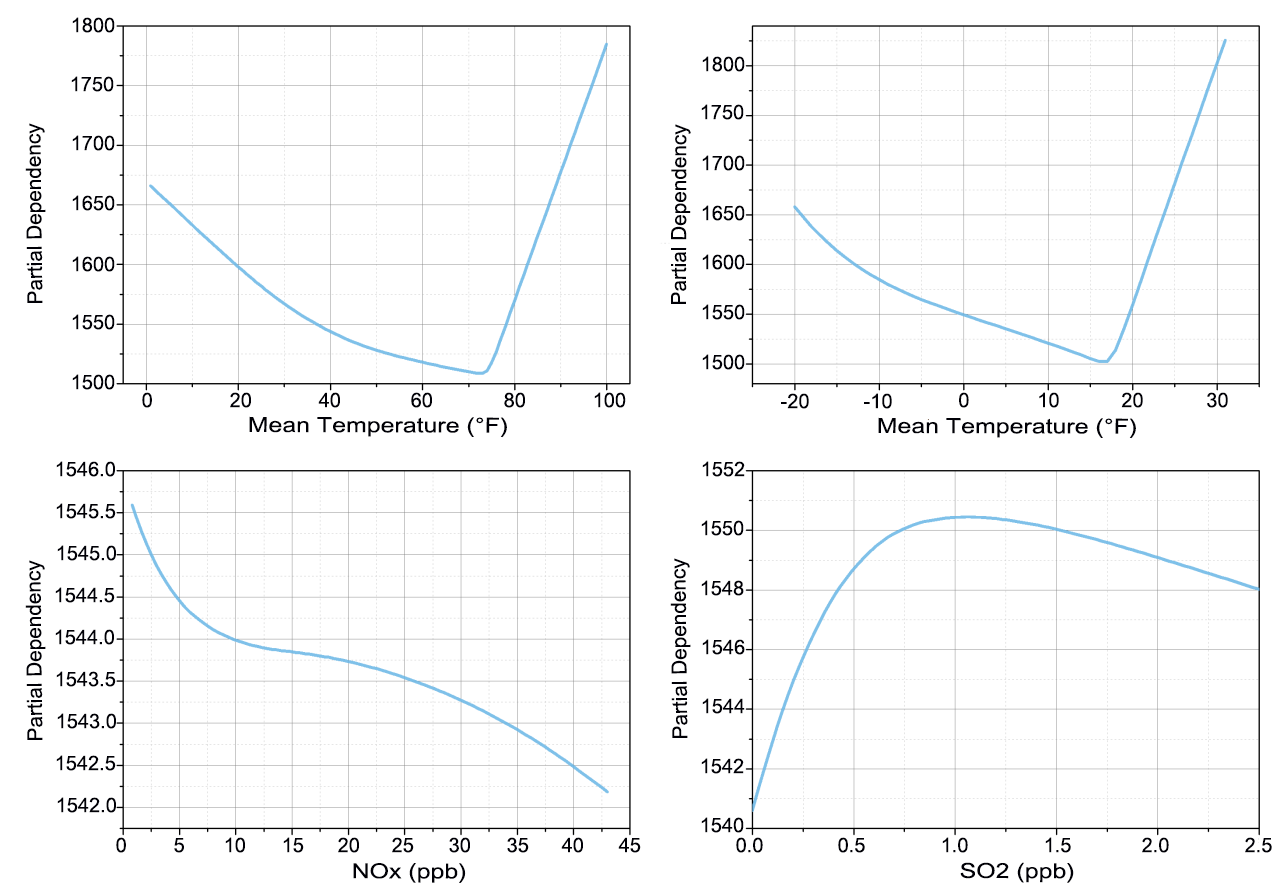}%
		\label{fig72}}
	\caption{Beesworm plots and partial dependency graphs of four geographical dimension features.}
	\label{fig7}
	
\end{figure*}

\begin{figure*}[!h]
	\centering
	\subfloat[]{\includegraphics[width=0.49\textwidth,height=0.45\textwidth]{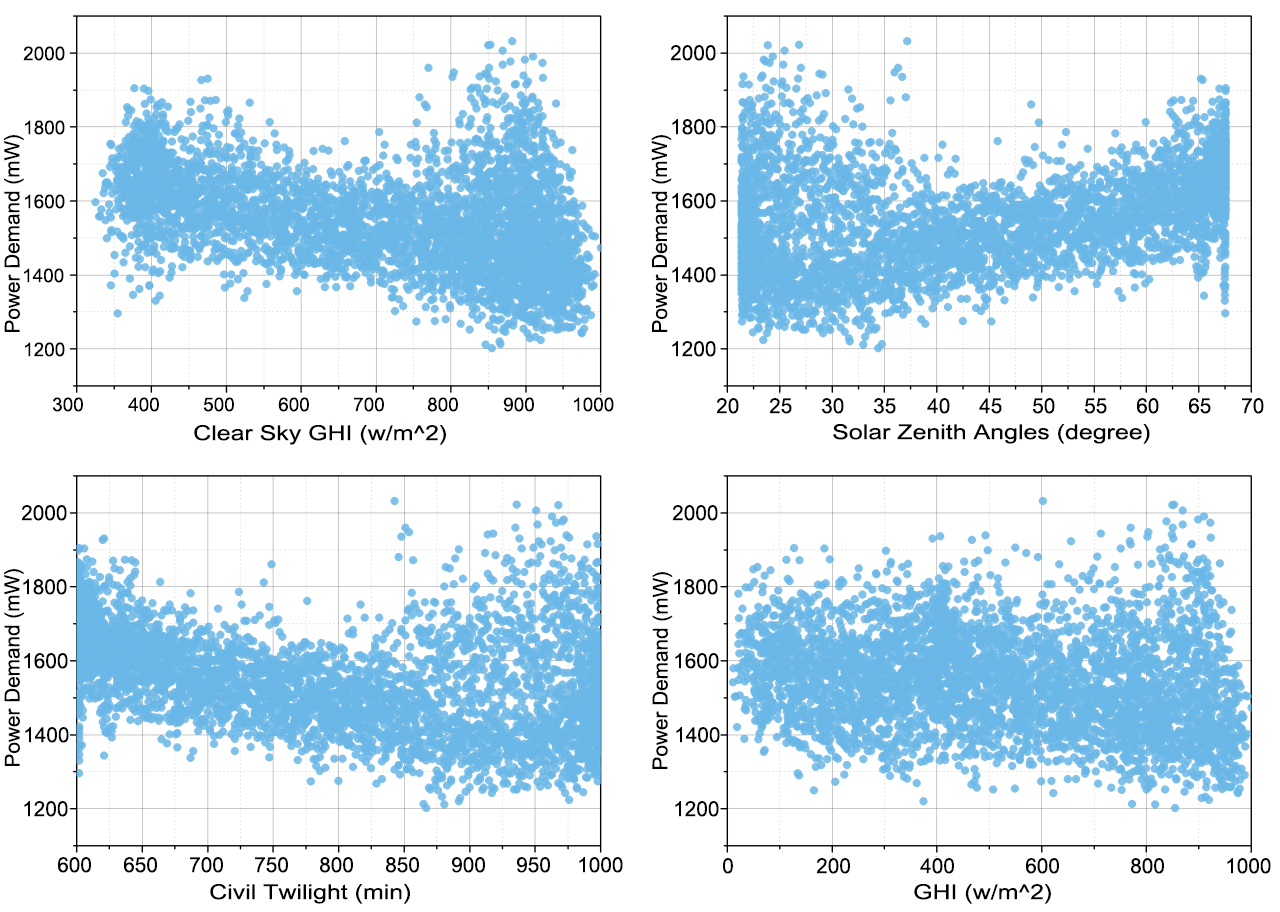}%
		\label{fig81}}
	\hfil
	\subfloat[]{\includegraphics[width=0.49\textwidth,height=0.45\textwidth]{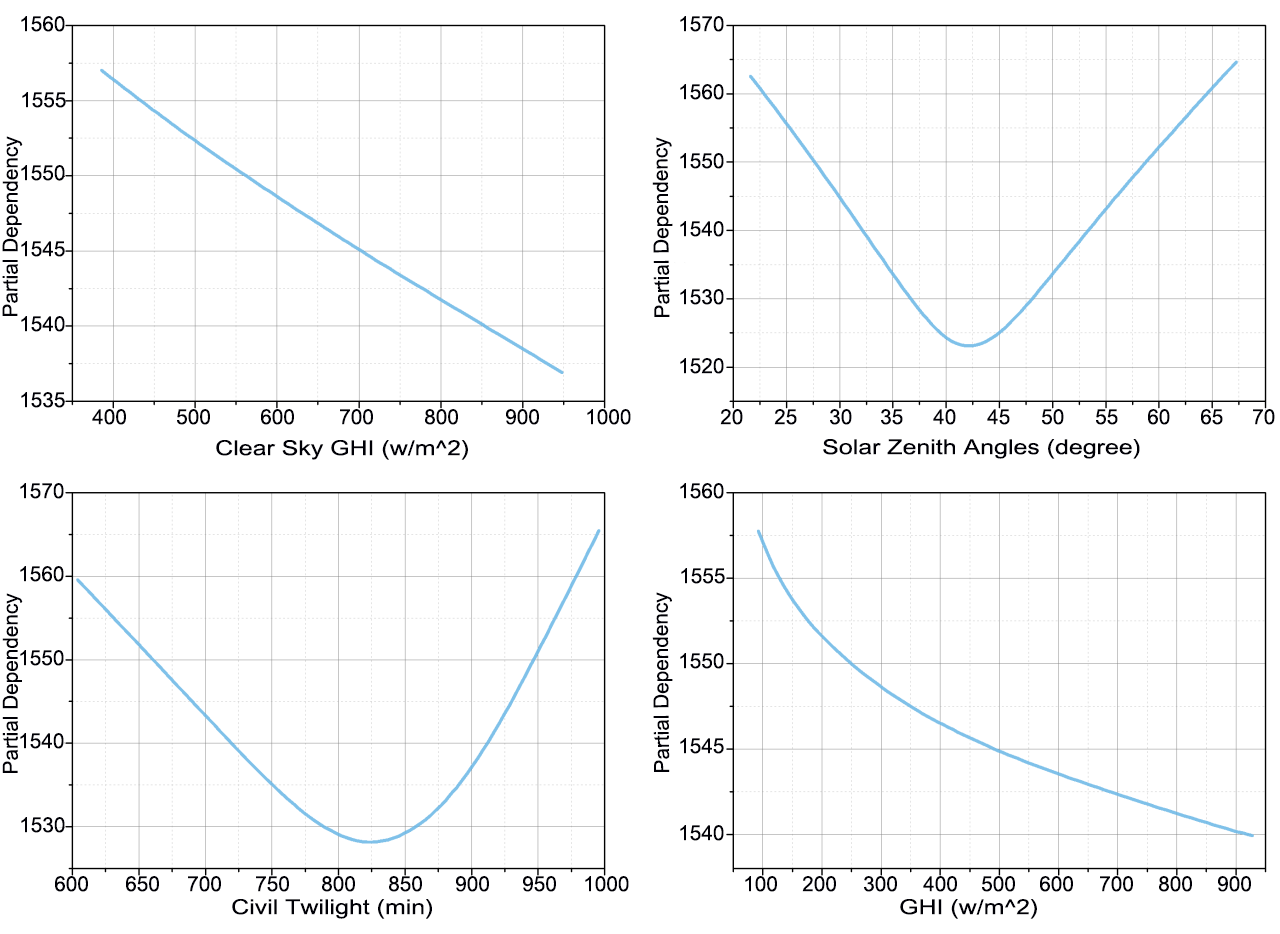}%
		\label{fig82}}
	\caption{Beesworm plots and partial dependency graphs of four astronomical dimension features.}
	\label{fig8}
\end{figure*}

\begin{figure*}[!h]
	\centering
	\subfloat[]{\includegraphics[width=0.3\textwidth,height=0.28\textwidth]{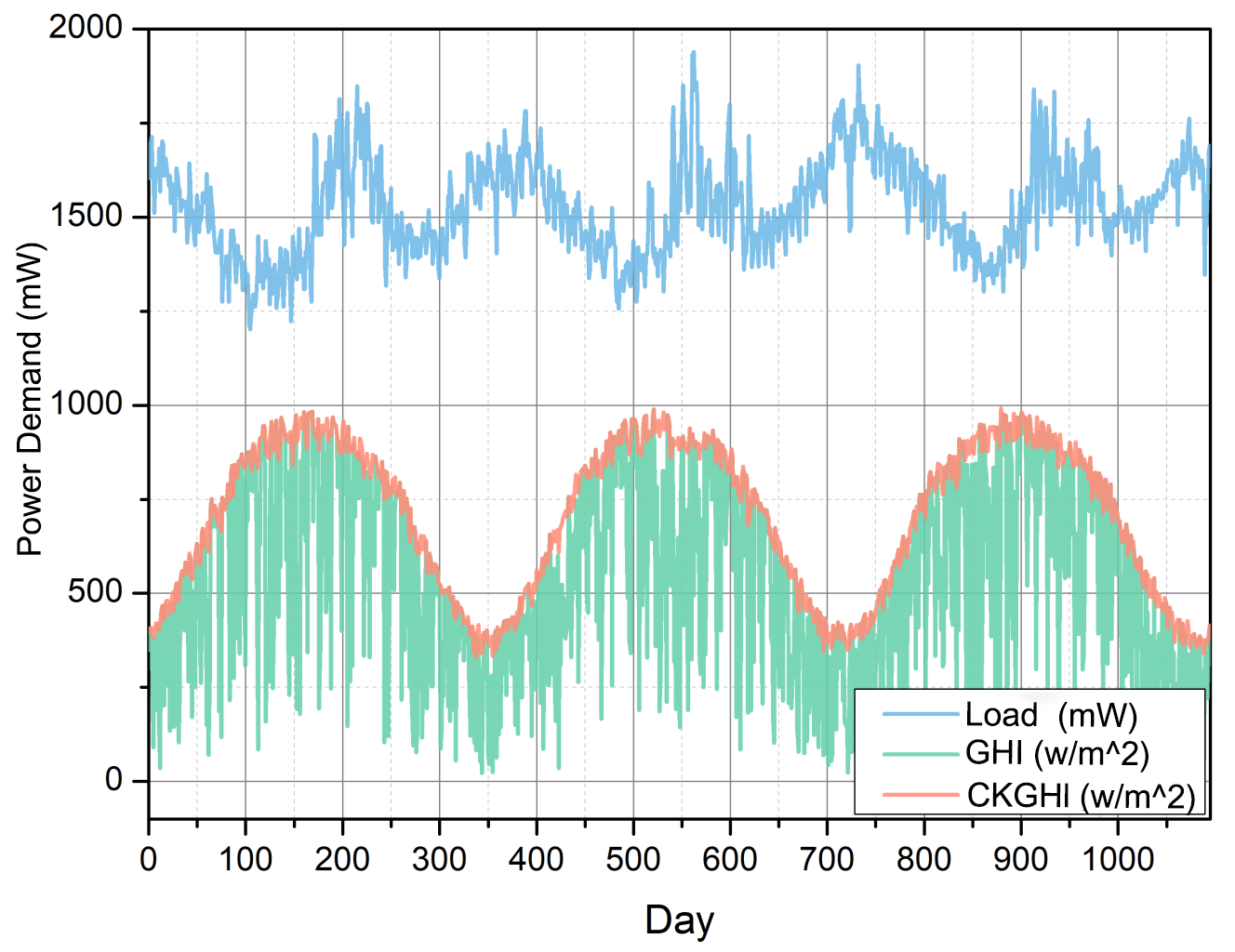}%
		\label{fig91}}
	\hfil
	\subfloat[]{\includegraphics[width=0.3\textwidth,height=0.28\textwidth]{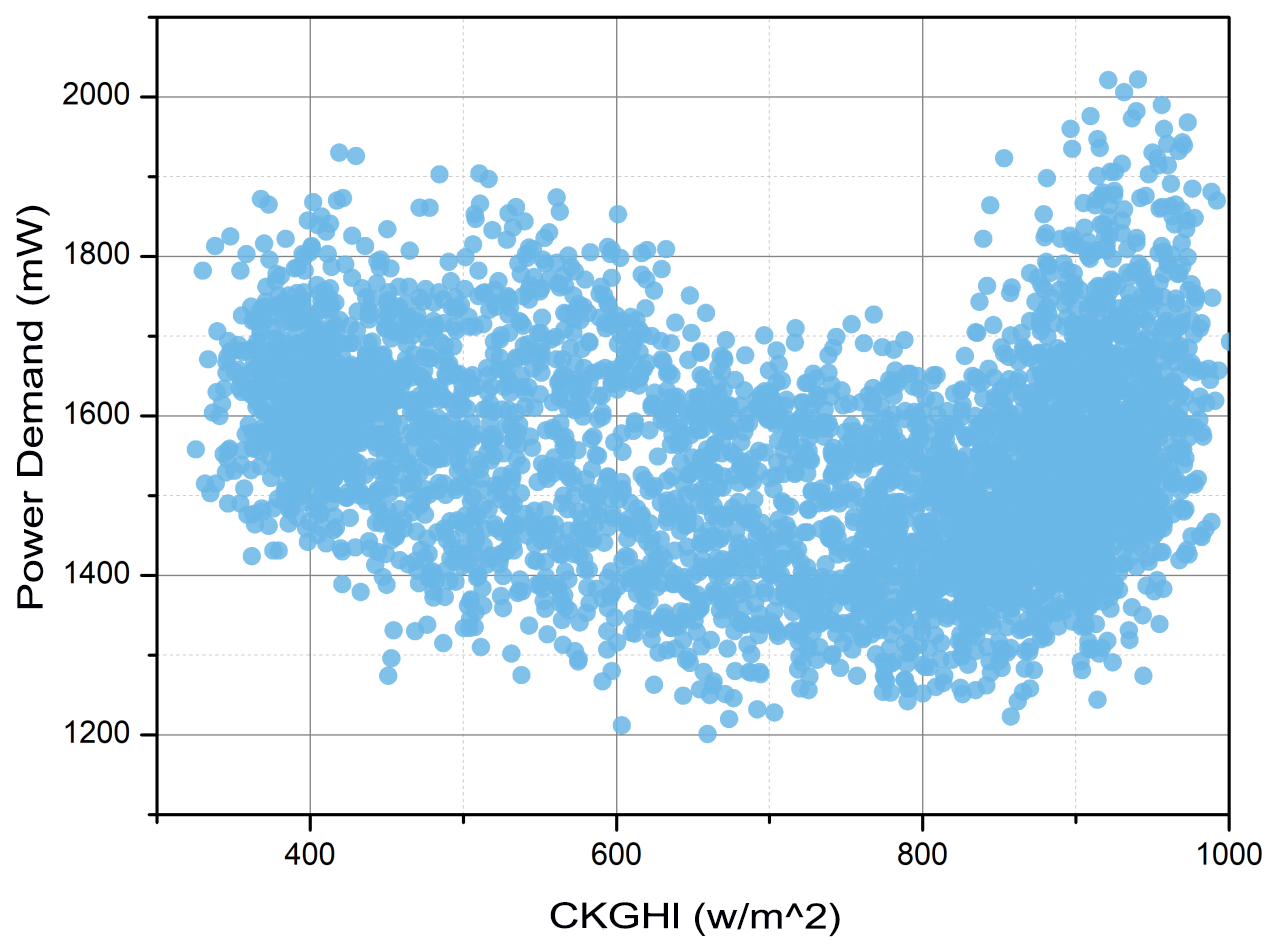}%
		\label{fig92}}
	\caption{Beesworm plots of load, GHI, and lag CKGHI in Maine.}
	\label{fig9}
\end{figure*}

\begin{figure*}[!h]
	\centering
	\subfloat[]{\includegraphics[width=0.78\textwidth,height=0.24\textwidth]{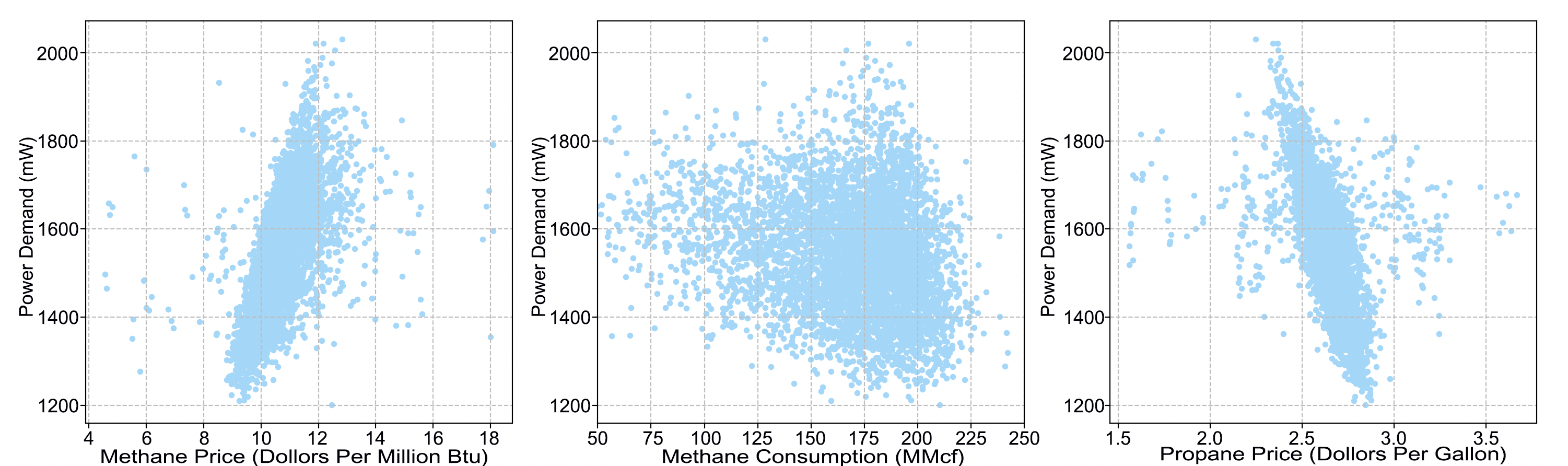}%
		\label{fig1222}}
	\hfil
	\subfloat[]{\includegraphics[width=0.75\textwidth,height=0.24\textwidth]{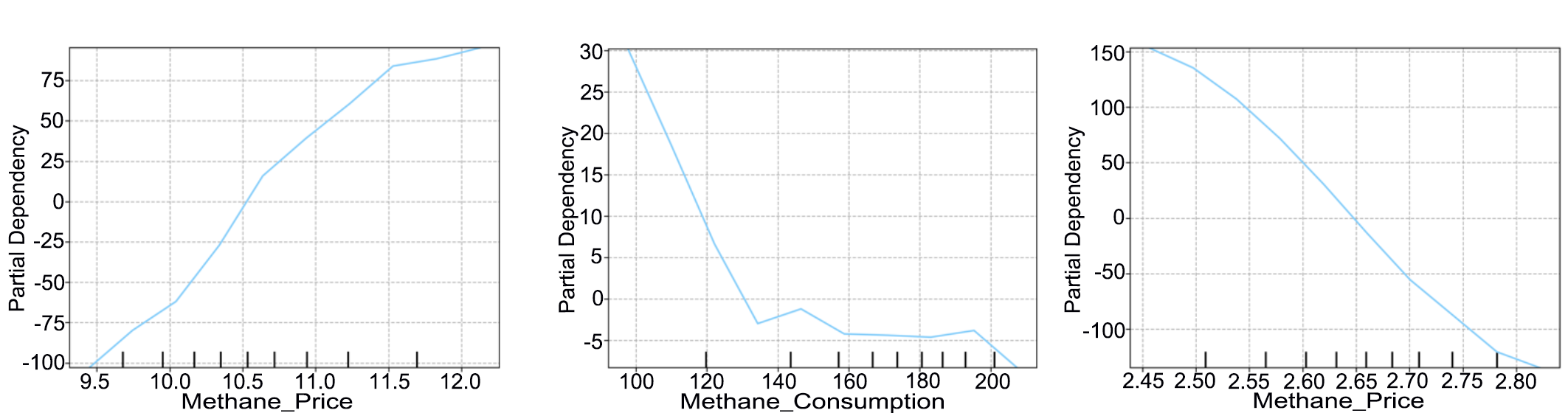}%
		\label{fig102}}
	\caption{Beesworm plots and partial dependency graphs of three integrated energy dimension features.}
	\label{fig10}
\end{figure*}

\subsection{\textit{Dependency correlation analysis}}
\hspace{16pt}To further investigate the correlation between features and demand load changes and how dominant features affect demand forecast results, this paper employs partial dependence graphs and beeswarm plots to explore the relationships between load fluctuation and dominant factors in the Maine case. For a trained model, the partial dependence graph can depict the response of the model's forecasting results to a single feature change. The partial dependence function is defined as follows.

\begin{equation}
	\hat{f(x_j)} = \frac{1}{n}\sum_{i=1}^{n}\hat{f(x_j, x_{-j}, i)}
\end{equation}

Wherein, $\hat{f}$ represents the trained model, n represents the number of samples in the training set, $x_{-j}$ represents other features except for $x_j$ , and the partial dependence of $x_j$ is defined as the mean value of forecasted values obtained by $\hat{f}$ force when $x_j$ is fixed, and $x_{-j}$ changes within its range.

We focus on the maximum and mean temperature features to examine the relationship between temperature and load, and NOx and SO2 to investigate the relationship between atmospheric indices and load. As shown in Figure 8(a), the fluctuation between the two temperature features and the load presents a V-shape. This phenomenon is also evident in Figure 8(b), where the change of temperature and load exhibits an inverse linear relationship below and above a certain temperature threshold, referred to as the temperature equilibrium point\cite{alipour2019assessing}. The equilibrium point is approximately 70°F for the maximum temperature and 15°F for the mean temperature. When the temperature exceeds or falls below this threshold, the use of air conditioning or heating equipment leads to an increase in power demand. In terms of the partial dependence graph of NOx, the forecasted load value and NOx content demonstrate a negative correlation. Moreover, based on the partial dependence graph of SO2, the load forecasted value increases when the SO2 content is between 0 and 1 ppb. However, when the content exceeds 1 ppb, the load forecasting value decreases slowly.

It can be seen from Figure 9(b) that for the solar zenith angles (SZA) and civil twilight duration, there is also a V-shaped relationship between them and the forecasted load value, so these two factors also have the equilibrium point when the load is at its lowest. It can be seen from the figure that the SZA equilibrium point is about 42°, and the civil dawn length equilibrium point is about 820 minutes. When latitude is determined, the position between the Earth and the Sun is reflected by the SZA at noon and the duration of civil dawn, and they have a significant annual periodic change rule. When the SZA and the civil dawn duration are in the intermediate value, the place is in spring or autumn, and when they are higher or lower, it is summer or winter. As shown in Figures 9(a) and 9(b), load fluctuations are negatively correlated with the GHI in the clear sky.

As shown in Figure 10 (a), the daily peak load, the global horizontal irradiance (GHI), and the clear sky global horizontal irradiance (CKGHI) all present annual periodic changes, but the change period of the GHI and the CKGHI are twice as long as the daily peak load. The two peaks of the load corresponding to the peak and trough of the CKGHI to a certain extent, but there is a specific phase difference between the two, and the load lags behind the CKGHI for about 50 days. It can be seen from Figure 10(b) that the CKGHI and the daily peak load show a V-shape after the lag of 50 days. Due to the large specific heat capacity of the ocean, the heat accumulation received by the Earth from the Sun lags to a certain extent relative to the solar radiation. The heat absorbed by the Earth will reach the maximum value only after 1 to 2 months have passed the maximum value of the solar radiation. This heat accumulation comprehensively leads to changes in climate and weather in a specific place and further influences load fluctuations. Therefore, load fluctuations have a certain lag in response to changes in solar radiation.

Figure 11 illustrates the relationship between power demand and three energy factors: methane price, methane consumption, and propane price. The methane price shows a positive linear correlation with power demand, while methane consumption and propane price show negative linear correlations with power demand. This relationship can be explained by the properties of these energy sources. Biogas, which is primarily represented by methane, forms a complementary relationship with power generation. Thus, an increase in methane prices leads to a corresponding increase in power consumption. On the other hand, propane, which is the primary energy source exported by the United States, competes with power generation for consumption. A rise in propane prices decreases export volume, promotes internal consumption, reduces power generation, and ultimately negatively affects power demand.

Further analysis discovered that certain features have a lagging effect on power demand. We selected several features with significant lagging effects for analysis. In terms of geographic dimensions, the daily average temperature remains stable with minimal fluctuations. However, the peak of the daily average temperature coincides with the peak of power demand, and has a lagging effect on power demand of around 10 days. The peaks and valleys of the daily average dew point temperature also coincide with power demand, and have a lagging effect on power demand of approximately 5-10 days. In terms of astronomical dimensions, sun-related features have a more significant lagging effect on power demand. These features are positively correlated with power demand as a whole, and have a lagging effect on power demand of approximately 50 days. Features in the IES dimension are relatively stable. Among them, natural gas consumption shows a clear seasonal fluctuation trend and is negatively correlated with power demand, with a lagging effect of approximately 30 days. Correspondingly, natural gas prices fluctuate positively with power demand and have a 30-day lagging effect on power demand. On the other hand, propane prices fluctuate negatively with power demand and have a 30-day lagging effect on power demand.

This study is carried out in Maine and Texas, U.S. According to the data provided by Wind company, the natural gas price in the United States and Europe reached new record highs in 2020-21. On September 27, 2021, U.S. NYMEX October natural gas futures closed up 11.01\%, setting a new record since 2014. With the soaring price of natural gas, the power demand in both the United States and Europe follows an obviously increased. This phenomenon validates the conclusion revealed by TKNFD that methane price acts as a dominant feature of power demand and positively correlates with power demand, which also strongly indicates the applicability of TKNFD to different regions.

\section{Conclusion}

\hspace{16pt}In this study, we propose a feature discovery method called TKNFD, a novel knowledge discovery method that actively acquire rich textual knowledge and discover numerical-modal features guided by textual knowledge without requiring power demand theory and big public databases. Specifically, TKNFD can extensively accumulate and cluster qualitative textual knowledge through web crawling and text mining. Under the guidance of the rich knowledge, TKNFD creates the candidate feature-type set and numerical 4DM-STD databases, which cover twice the domain dimension and 5-9 times the number of candidate features than empirical methods. Finally, TKNFD can quantitatively identify features by a hierarchical and model-independent strategy, and systematically analyses the contribution of different features and dependency correlation between features and power demand. Benchmark forecast experiments using five classical and advanced forecasting models in two different regions demonstrate that TKNFD-discovered 43-48 dominant features outperform the state-of-the-art feature schemes by 16.84\% to 36.36\% MAPE. These verify that out proposed TKNFD could reliably discover more comprehensive features. Moreover, these confirm that the features we have discovered could improve the forecasting accuracy of power demand by at least 16.84\% MAPE, regardless of the types of forecasting models.

In particular, TKNFD reveals that more features exist in the empirically known geographic and social dimensions, and many features, especially several dominant features, exist in the unknown integrated energy and astronomical dimensions. The contribution of each dimension and feature are concluded as follows. (1) The integrated energy dimension has the most significant improvement on SPDF accuracy, in which the methane price is the dominant feature and is approximately 30 days lagging positively correlated with power demand. (2) In astronomical dimension, the sun-related features play important roles and generally cast a 50 days lagging effect on power demand. Solar zenith angle, civil twilight duration, and lagged clear sky global horizontal irradiance have a V-shape relationship with power load, indicating that there exist balance points for them. Global horizontal irradiance is negatively related to power demand. (3) In the geographical dimension, temperature is an important feature affecting the load change, which also shows a V-shape relationship with power demand. (4) The social dimension has the weakest impact among four dimensions. Saturday and Monday are more important than other features in this dimension. 

These findings deepen our understanding of the complex and nonlinear nature of power demand fluctuations in the ongoing low-carbon transition of power sectors and IES. By shedding light on the contribution of different dimensions and features, this study offers valuable insights for policymakers and planners seeking to promote collaboration and competition between the power sector and other energy sectors in constructing new IES with low-carbon goals.

In addition, we have made much effort to build two 4DMSTDs including four feature dimensions and 58-92 features. As a matter of fact, we have qualitatively found a large number of features in the candidate feature-type set. However, due to the current unavailability of numerical data for some of these features, these features are not included in 4TD-SDM. Despite the imperfect completeness of 4DM-STDs, they can serve as public baseline feature databases for SPDF .

The proposed TKNFD is not limited to feature discovery for power demand forecasting in power sector. It is applicable to feature discovery in other sectors such as energy system, epidemic forecasting tasks where the mechanism remains unclear and public feature databases are unavailable.



\bibliographystyle{IEEEtran}
\bibliography{refs2023}
%
%
%
%

\end{sloppy}
\end{document}